\pdfoutput=1
\documentclass[runningheads]{llncs}
\usepackage{graphicx}
\usepackage{comment}
\usepackage{amsmath,amssymb} 
\usepackage{color}
\usepackage{multirow}
\usepackage{hyperref}

\begin{document}
\pagestyle{headings}
\mainmatter
\def\ECCVSubNumber{21}  

\title{Deformable PV-RCNN: Improving 3D Object Detection with Learned Deformations} 

\institute{Paper ID \ECCVSubNumber}

\author{Prarthana Bhattacharyya \and
Krzysztof Czarnecki}
\institute{University of Waterloo \\
\email{\{p6bhatta, k2czarne\}@uwaterloo.ca}}
\titlerunning{Deformable PV-RCNN: Improving 3D Object Detection}
\maketitle

\begin{abstract}
We present Deformable PV-RCNN, a high-performing point-cloud based 3D object detector. Currently, the proposal refinement methods used by the state-of-the-art two-stage detectors cannot adequately accommodate differing object scales, varying point-cloud density, part-deformation and clutter. We present a proposal refinement module inspired by 2D deformable convolution networks that can adaptively gather instance-specific features from locations where informative content exists. We also propose a simple context gating mechanism which allows the keypoints to select relevant context information for the refinement stage. We show state-of-the-art results on the KITTI dataset.
\end{abstract}

\section{Introduction}
3D object detection from point clouds is critical for autonomous driving and robotics. We build on the success of PV-RCNN \cite{PVRCNN}, a state-of-the-art 3D object detector. 
\\ \\ \textbf{Motivation} Part of PV-RCNN’s success is due to the randomly sampled keypoints which capture multi-scale features for proposal refinement while retaining fine-grained localization information. However, random sampling is not effective over potentially ambiguous scenes. For example, ‘pedestrians’ and ‘traffic poles’ can be very hard to distinguish in point clouds. In this case, we wish to align the keypoints towards the most discriminative areas, so that principal features for a pedestrian can be highlighted. Similarly, the scales for cars, pedestrians and cyclists are very different. While multi-scale feature aggregation is advantageous for image features, the non-uniform density of point clouds makes it hard to detect them with a single model. We wish to adaptively aggregate and focus on their most salient features at various scales. Lastly to handle clutter and avoid false positives, for example, to avoid detecting all seated individuals as cyclists, we need to pick up on unevenly distributed contextual information. 
\\ \\ \textbf{Contribution} We construct Deformable PV-RCNN, a 3D detector that handles sparsity of LIDAR points, is adaptive to non-uniform point cloud density especially at long distances and can address clutter in real-world traffic scenes. We show that we can outperform PV-RCNN across different categories and especially at long-distances on the KITTI 3D object detection dataset.

\begin{figure}[t]
\centering
\includegraphics[height=2.8cm, width=8cm]{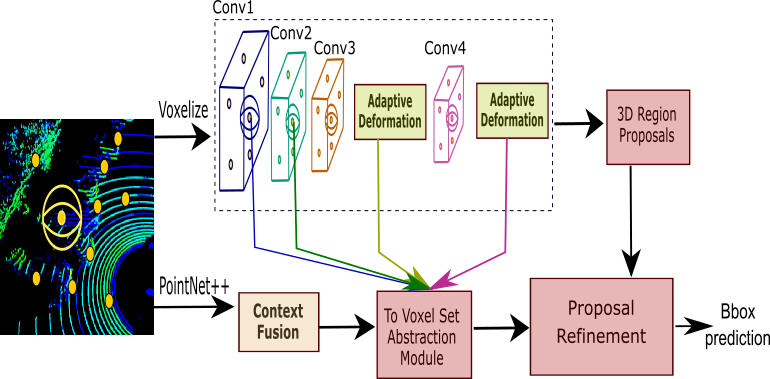}
\caption{Our architecture for Deformable PV-RCNN. We propose an Adaptive Deformation and Context Fusion module.}
\label{fig:full}
\end{figure}

\begin{figure}[t]
\centering
\includegraphics[height=2.2cm, width=7cm]{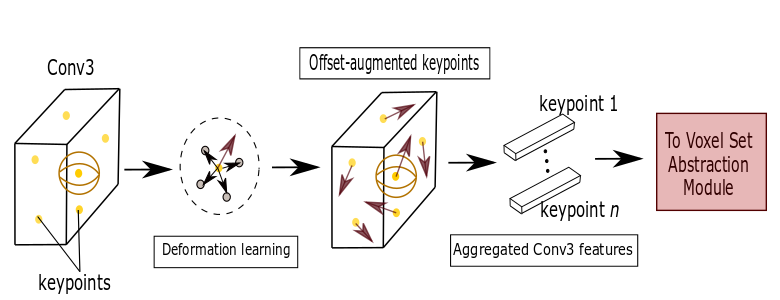}
\caption{Adaptive deformation module adaptively aligns keypoints towards the most-content rich and discriminative features.}
\label{fig:def}
\end{figure}

\section{Methods}
Our 3D detection pipeline is presented in Fig.~1. It consists of an Adaptive Deformation module (Fig.~2) and a Context Fusion module (Fig.~3).
\begin{figure}[b]
\centering
\includegraphics[height=1.6cm, width=7cm]{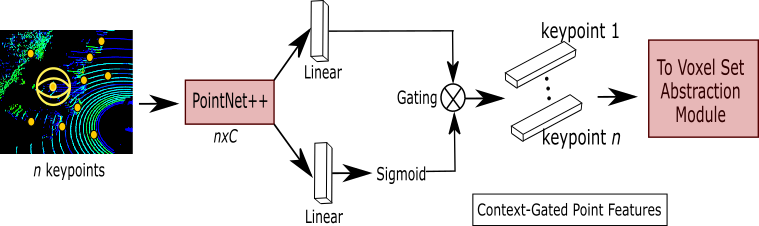}
\caption{Context fusion module dynamically selects most relevant context features.}
\label{fig:context}
\end{figure}

\textbf{Adaptive Deformation} The $n$ sampled keypoints (shown in yellow in Fig.~1) have a 3D position $v_i$ and a feature vector $f_i$ corresponding to either of Conv3 or Conv4 layers. Our module computes updated features $f'_i$ as follows: \\
$f'_i = \frac{1}{n}$ReLU$(\sum_{j \in \mathcal{N}(i)}{W_\textit{offset}(f_i-f_j)\cdot(v_i-v_j)})$, where $\mathcal{N}_i$ gives the $i$-th keypoint's neighbors in the point-cloud and $W_\textit{offset}$ is a learned weight matrix. We then obtain the new deformed keypoint positions as $v'_i = v_i+$tanh$(W_\textit{align}[f'_i])$, where $W_\textit{align}$ is a learned weight matrix. This is similar to \cite{MeshRCNN}, \cite{Pointdan}. We then proceed to compute the features for the deformed keypoints using PointNet++ similar to the PV-RCNN pipeline.

\textbf{Context Fusion}
This module uses context gating to dynamically select representative and discriminative features from local evidence, highlighting object features and suppressing clutter. Given a keypoint feature $f_i$, the modulating feature is obtained as $g = \sigma(W_\textit{gate} f_i + b_\textit{gate})$ and the context-gated feature is computed as $f^g_i = g \odot W_\textit{fc}f_i$, where $W_\textit{gate}$, $b_\textit{gate}$, $W_\textit{fc}$ are learned from data \cite{GLU}.

\section{Results}
\setlength{\tabcolsep}{4pt}
\begin{table}[t]
\begin{center}
\caption{Performance comparison on the moderate level of KITTI \textit{val} split with AP calculated calculated by 11 recall positions}
\label{table:headings}
\begin{tabular}{c||ccc}
\hline
Model & Car & Cyclist & Pedestrian\\
\hline
SECOND \cite{SECOND}  & 78.62 & 67.75 & 52.98 \\
F-PointNet \cite{FrustrumPointnet} & 70.92 & 56.49 & \bf{61.32} \\
Part-A2 \cite{PartA2}  & 79.40 & 69.90 & 60.05 \\
PV-RCNN \cite{PVRCNN} & \bf{83.69} & 69.47 & 54.84 \\
Ours  & \bf{83.30} & \bf{73.46} & 58.33 \\
\hline
\end{tabular}
\end{center}
\end{table}
\setlength{\tabcolsep}{1.4pt}

\setlength{\tabcolsep}{4pt}
\begin{table}[t]
\begin{center}
\caption{Ablation study on the moderate level of KITTI \textit{val} split with AP calculated calculated by 40 recall positions}
\label{table:headings}
\begin{tabular}{cc||ccc}
\hline
Deformations & Context-Fusion & Car & Cyclist & Pedestrian\\
\hline
&  & 84.20 & 69.65 & 54.49 \\
\checkmark & & 84.24 & 70.21 & 57.31 \\
\checkmark & \checkmark & \bf{84.71} & \bf{73.03} & \bf{57.65} \\
\hline
\end{tabular}
\end{center}
\end{table}
\setlength{\tabcolsep}{1.4pt}

\setlength{\tabcolsep}{4pt}
\begin{table}[t]
\begin{center}
\caption{Comparison of nearby and distant-object detection on the moderate level of KITTI \textit{val} split with AP calculated by 40 recall positions}
\label{table:headings}
\begin{tabular}{c|c|ccc}
\hline
Distance & Model & Car & Cyclist & Pedestrian\\
\hline
 \multirow{2}{*}{0-30\,m} & PV-RCNN & \bf{91.71} & 73.76 & 56.82 \\ & Ours & 91.65 & \bf{74.89} & \bf{59.61} \\
 \hline
 \multirow{2}{*}{30-50\,m} & PV-RCNN & 50.00 & 35.15 & - \\ & Ours & \bf{52.02} & \bf{47.00} & - \\
\hline
\end{tabular}
\end{center}
\end{table}
\setlength{\tabcolsep}{1.4pt}

The 3D object detection benchmark of KITTI \cite{KITTI} contains 7481 training samples. Following \cite{MV3D}, we divide them into 3712 training samples and 3769 validation samples. We report our results on the validation (\textit{val}) split of KITTI. \\ \\
\textbf{Comparison with state-of-the-art methods}  Table~1 shows the performance of Deformable PV-RCNN on the KITTI \textit{val} split. We run the officially released checkpoints for SECOND, Part-A2 and PV-RCNN.\footnote{\url{https://github.com/open-mmlab/OpenPCDet}} Our method outperforms the state-of-the-art on the \textit{cyclist} class. Compared to PV-RCNN, it performs better on \textit{pedestrian} and \textit{cyclist} class by 3.5 \% and 4\%, respectively, and on par for the \textit{car} class. \\ \\
\textbf{Ablation Studies} Table~2 validates the effectiveness of learning deformable offsets and context gating by comparing with a retrained PV-RCNN baseline. We find that deformation prediction contributes to the performance increase over all classes, but especially for the \textit{pedestrian} class, which is probably due to being able to focus on representative features like arms and legs. Context fusion also increases performance over all classes, indicating that contextual information is important for refinement. \\ \\
\textbf{Comparison with PV-RCNN} Table~3 shows that our model outperforms the baseline at different distances for the \textit{pedestrian} and \textit{cyclist} class. We also find a 2\% increase in AP for the \textit{car} class at long distances where the scans are sparse and context information becomes important.  Fig.~4 shows that similar AP performance can be obtained with fewer keypoints as compared to the baseline across all classes, which is probably due to their deformability to cover representative locations. Fig.~5 shows qualitative 3D detection results.

\section{Conclusion}
We present Deformable PV-RCNN, a 3D object detector for detecting 3D objects from raw point cloud.  Our proposed deformation prediction method adaptively aligns the keypoints encoding the scene towards the most discriminative and representative features. The proposed context gating network adaptively highlights relevant context features thereby favouring more accurate proposal refinement. Our experiments show that Deformable PV-RCNN outperforms PV-RCNN in various challenging cases on the 3D detection benchmark of KITTI dataset, especially benefiting smaller objects and complex scenes more.

\begin{figure}[t]
\centering
\includegraphics[height=2.5cm, width=10.5cm]{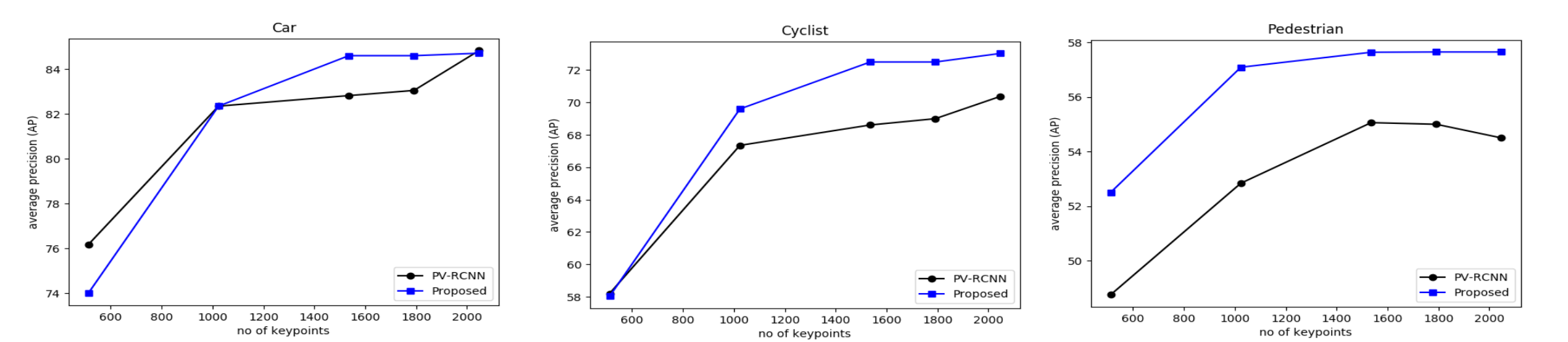}
\caption{AP of PV-RCNN (black) vs. Deformable PV-RCNN (blue) for a number of keypoints, varying from 512 to 2048}
\label{fig:graph}
\end{figure}

\begin{figure}[h]
\centering
\includegraphics[height=2cm, width=\textwidth]{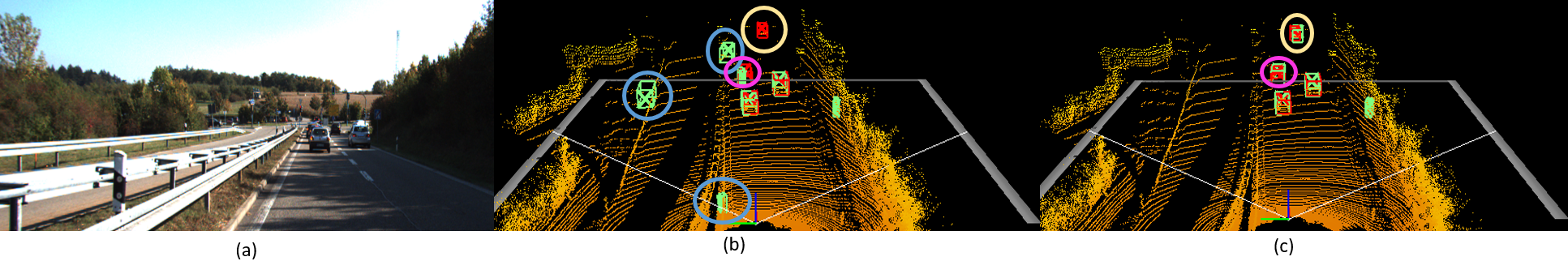}
\caption{From left to right: RGB-image; PV-RCNN output; Deformable PV-RCNN output. GT boxes visualized in red and predictions in green. Yellow and pink circled objects are picked up by our detector but missed or oriented wrongly by PV-RCNN; Objects circled blue are clutter which our detector suppressed.}
\label{fig:qual}
\end{figure}

\clearpage
%
%
\bibliographystyle{splncs04}
\bibliography{defpvrcnn.bib}

\end{document}